\documentclass{article}
\usepackage{spconf,amsmath,graphicx}

\usepackage{algorithm} 
\usepackage[noend]{algpseudocode}

\usepackage{enumitem} 

\title{
ORGB: OFFSET CORRECTION IN RGB COLOR SPACE
\\ FOR ILLUMINATION-ROBUST IMAGE PROCESSING
}
\name{Zhenqiang Ying$^1$, Ge Li$^1$ 
\thanks{This work was supported by the grant of National Science Foundation of China (No.U1611461), Shenzhen Peacock Plan (20130408-183003656), and Science and Technology Planning Project of Guangdong Province, China (No. 2014B090910001).}, 
Sixin Wen$^2$, Guozhen Tan$^2$} 
\address{$^1$SECE, Shenzhen Graduate School, Peking University, Shenzhen, China\\
$^2$Faculty of Electronic Information and Electrical Engineering, Dalian University of Technology, Dalian, China\\
}

\begin{document}
%
\maketitle
\begin{abstract}
Single materials have colors which form straight lines in RGB space.
However, 
in severe shadow cases,
those lines do not intersect the origin, which is inconsistent with the description of most literature.
This paper is concerned with the detection and correction of the offset between the intersection and origin.  
First, we analyze the reason for forming that offset via an optical imaging model.
Second, we present a simple and effective way to detect and remove the offset.
The resulting images, named ORGB,
have almost the same appearance as the original RGB images
while are more illumination-robust 
for color space conversion.
Besides, image processing using ORGB instead of RGB is free from the interference of shadows.
Finally, the proposed offset correction method is applied to road detection task,
improving the performance both in quantitative and qualitative evaluations.


\end{abstract}
\section{Introduction}
\label{sec:intro}



Non-uniform illumination confounds many computer vision algorithms. 
In particular, shadows in an image can lead segmentation, tracking, or recognition algorithms to fail. 
The reason lies in that 
the illumination-sensitive brightness and illumination-insensitive chromaticity 
information are mixed together in all three components of an RGB image.
To remove the interference of non-uniform illumination, 
Color Space Conversion is often used in the pre-processing stage 
to separate brightness from chromaticity information.
However, commonly used chromaticity spaces that claim to be illumination invariant 
are unstable in many natural situations, especially in the case of severe shadow.
The underlying reason is related to the 
inaccurate description
of pixel distribution characteristics.
It is proved via an ideal physical model that
the measured colors of homogeneous dielectric surfaces lie on a line passing through the origin of RGB space \cite{healey1992segmenting}.
Therefore, material surfaces will form straight lines that intersect the origin.
Unfortunately, 
it is not applicable in severe shadow cases, where
Omer and Werman \cite{omer2004color} observe that those lines are offset from the origin and do not intersect the origin if extended. 
The offset 
is also reported in \cite{maxwell2008bi}, from which a novel reflection model is proposed.
Taking the offset into consideration 
not only provides better image understanding but also helps   
improve the performance in many computer vision applications
such as road detection \cite{ying2016mm} and background subtraction \cite{huang2009moving}.

\begin{figure} [t]
\centering
\includegraphics[height=1.7in]{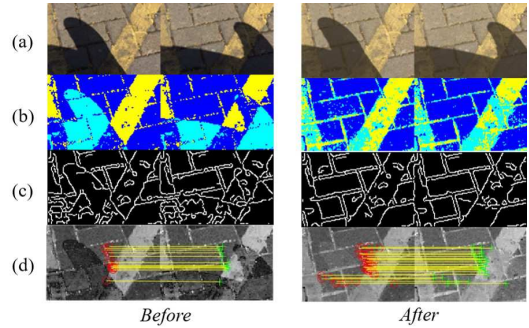}
\caption{
Image processing before and after offset-correction.
(a) Raw images (RGB and ORGB).
(b) Image segmentation using hue-saturation and k-means clustering.
(c) Canny edge detection on saturation.
(d) FAST feature match on saturation.
}
\label{fig-intro}
\end{figure}

While existing literature either extend the existing models or adopt a new algorithm,
this paper presents another perspective of dealing with the offset.
Instead of changing models or algorithms,
we try to change the images with the offset to that without the offset.
We name that process as offset correction. 
Using offset correction as a pre-processing step, 
existing techniques including Color Space Conversion can still be applied in severe shadow cases without
any modification.
After offset correction, commonly used chromaticity spaces like HSV, CIELUV are
illumination-robust even under severe shadow conditions.
Experiments show that 
the offset-corrected RGB images,
named ORGB, 
is more illumination-robust for image processing compared with the original RGB images,
as shown in Fig.\ref{fig-intro}. 
Besides, by applying the proposed offset correction 
into road detection task, the performance 
 is greatly improved
both in quantitative and qualitative evaluations.

\section{Related Work}
\label{sec:relatedWork}

Most color images are captured in RGB, 
but this color space is rarely used for computer vision tasks.
%
The reason is two-fold: on the one hand, the red, green and blue  components are highly correlated,
thus it is inefficient to process all components by analyzing nearly
the same picture three times. 
On the other hand, since the brightness and chromaticity information are mixed together for all three components of RGB color space, 
they are vulnerable to the impact of illumination effects. 
%
Many computer vision algorithms start with 
Color Space Conversion \cite{ying2016mm,Zhenqiang2016,Zhenqiang2015}, 
transforming RGB into color spaces where chromaticity components are separated from the brightness.


Although chromaticity components, e.g. hue and saturation, are illumination-robust against weak shadows,
they become very unstable in severe shadow conditions \cite{ying2016mm}. 
The reason lies in the aforementioned offset, which undermines 
the pixel distribution characteristics. 
Under the traditional assumption, single materials have colors 
which form straight lines intersect the origin in RGB space. 
Therefore,
as the intensity of the light changes, 
the brightness of the object changes while its color remains constant.
However, due to the existence of the offset between intersection and origin,
the object color do change in shadows \cite{maxwell2008bi,finlayson2006removal}.
Shafer \cite{shafer1985using} attributes the color inconstancy to that
the dynamic range and noise levels of cameras limit 
the utility of color measurements in shadows.
However, that inconstancy still exists in modern cameras 
whose dynamic range is sufficient to 
capture significant information in shadows \cite{maxwell2008bi}.
Therefore, the color inconstancy is physically present 
rather than be produced by cameras.


To make Color Space Conversion applicable in severe shadow cases,
we need to recover the color constancy. 
Since the offset causes color inconstancy, the recovery of color constancy
can be performed by detecting and removing that offset.
%
To find an effective way to perform offset correction,
we need to understand the reason for offset formation. 
In the next section,
We explain it through physical modeling and experimental verification.

\section{Offset Formation}
\label{sec:offsetFormation}



\begin{figure}[t]
\centering
\includegraphics[width=3.3in]{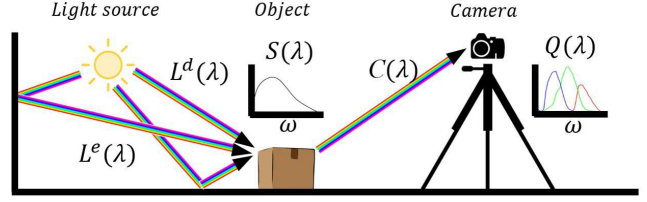}
\caption{Schematic diagram of image formation.} 
\label{fig:imageformation}
\end{figure}

To understand the formation of the offset, 
we take one pixel in an RGB image to analyze the imaging process.
A pixel can be represented as a triplet of numbers.
Denote $\boldsymbol{\rho} = [\rho_1,\rho_2,\rho_3]$ as the measured RGB colors in that pixel.
Generally, imaging devices sample the incoming light using three sensors, preferentially 
sensitive to long (red), medium (green), and short (blue) wavelength light respectively. 
The $k$-th measured color $\rho_k$ comes from the $k$-th sensor response, which can be modeled as
\begin{equation}
\label{eqn_rho}
\rho_k = \int_\omega C (\lambda) Q_k(\lambda) d\lambda, \quad k = 1, 2, 3,
\end{equation} 
where $Q_k(\lambda)$ is a function of wavelength $\lambda$ characterizing how sensor $k$ responds to the captured light $C(\lambda)$.
For matte surfaces 
that do not emit light, the captured light comes from reflection as
%
%
\begin{equation}
\label{eqn_c}
C (\lambda) = L(\lambda) S(\lambda),
\end{equation} 
where $L(\lambda)$ denotes the light incident upon a surface and $S(\lambda)$ is its surface reflectance function.
%
$L(\lambda)$ is composed of
direct light $L^d(\lambda)$ and environment light $L^e(\lambda)$.
The former comes directly from the light source while the latter comes from 
reflections of surrounding surfaces,  
as shown in Fig.\ref{fig:imageformation}.
The incident light can be modeled as
\begin{equation}
\label{eqn_l}
L (\lambda) = \mu L^d(\lambda) \cos\theta + L^e(\lambda),
\end{equation}
where  $\mu$ is a value between [0, 1] indicating how much direct light gets to the surface
and $\theta$ is the angle between the direct lighting and the surface norm.
Here $\mu$ is influenced by object occlusion and light attenuation.
Under outdoor daylight scenes where the attenuation of sun is negligible,
$\mu$ is 0 for umbra area, 1 for lit area and others for penumbra area.

Based on the above analysis, the overall sensor response 
comes from two type of light.
Define the contribution of direct light to the sensor response as 
%
\begin{equation} 
\phi_k = \mu \cos\theta \int_\omega L^d(\lambda) S(\lambda) Q_k(\lambda) d\lambda,  
\quad k = 1, 2, 3,
\end{equation}
and that of environment light as
\begin{equation}
\delta_k = \int_\omega L^e(\lambda) S(\lambda) Q_k(\lambda) d\lambda,  
\quad k = 1, 2, 3,
\end{equation}
then the overall sensor response is their summation:
\begin{equation}
\label{eqn_rho2} 
\boldsymbol{\rho} = \boldsymbol{\phi} + \boldsymbol{\delta},
\quad 
\boldsymbol{\phi} = [\phi_1,\phi_2,\phi_3], \boldsymbol{\delta} = [\delta_1,\delta_2,\delta_3].
\end{equation}

%
%

 

\begin{figure} [b]
\begin{minipage}[b]{0.31\linewidth}
  \centering
  \centerline{\includegraphics[height=1.3cm]{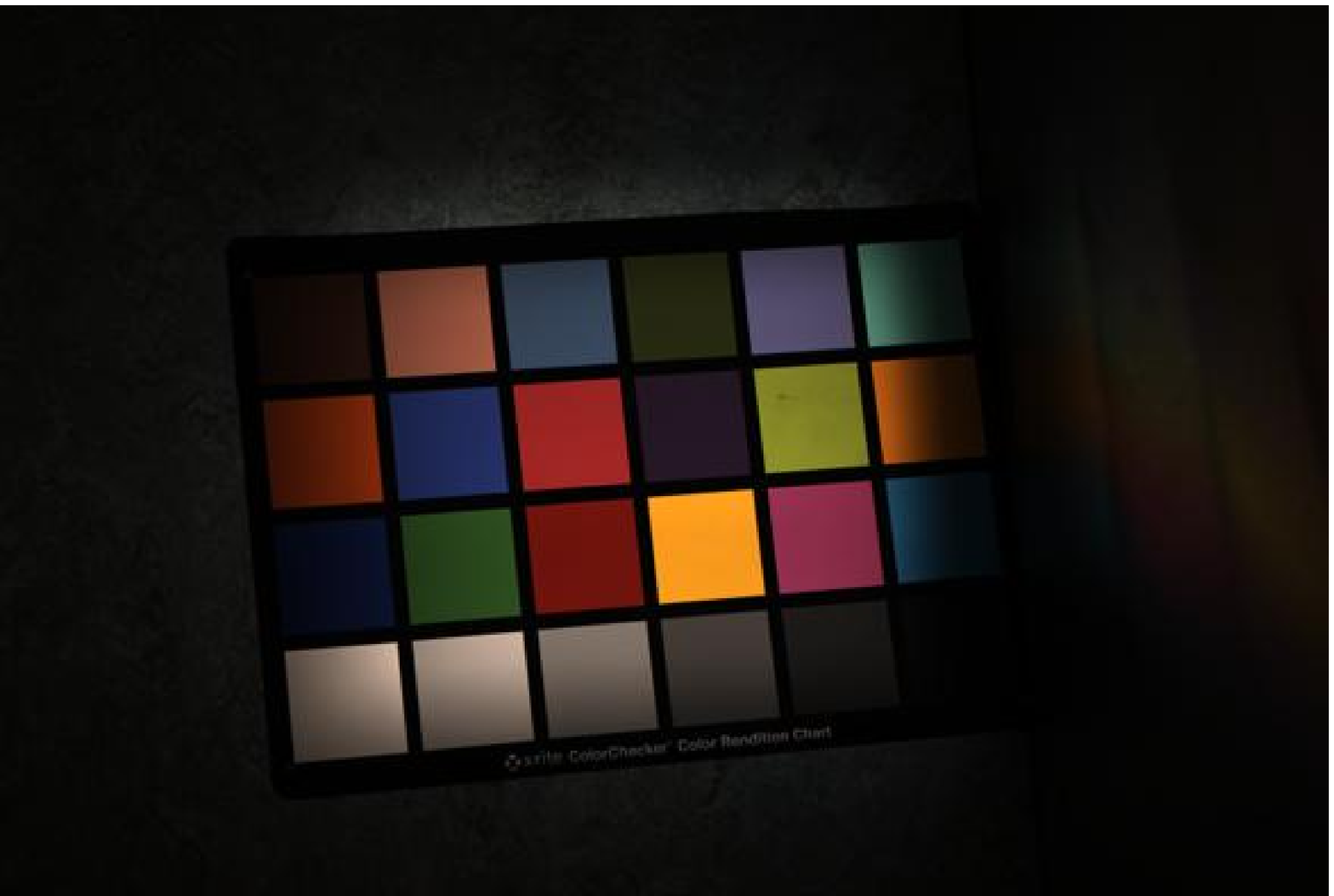}}
  \centerline{(a) $I_{\phi}$}\medskip 
\end{minipage}
\begin{minipage}[b]{0.31\linewidth}
  \centering
  \centerline{\includegraphics[height=1.3cm]{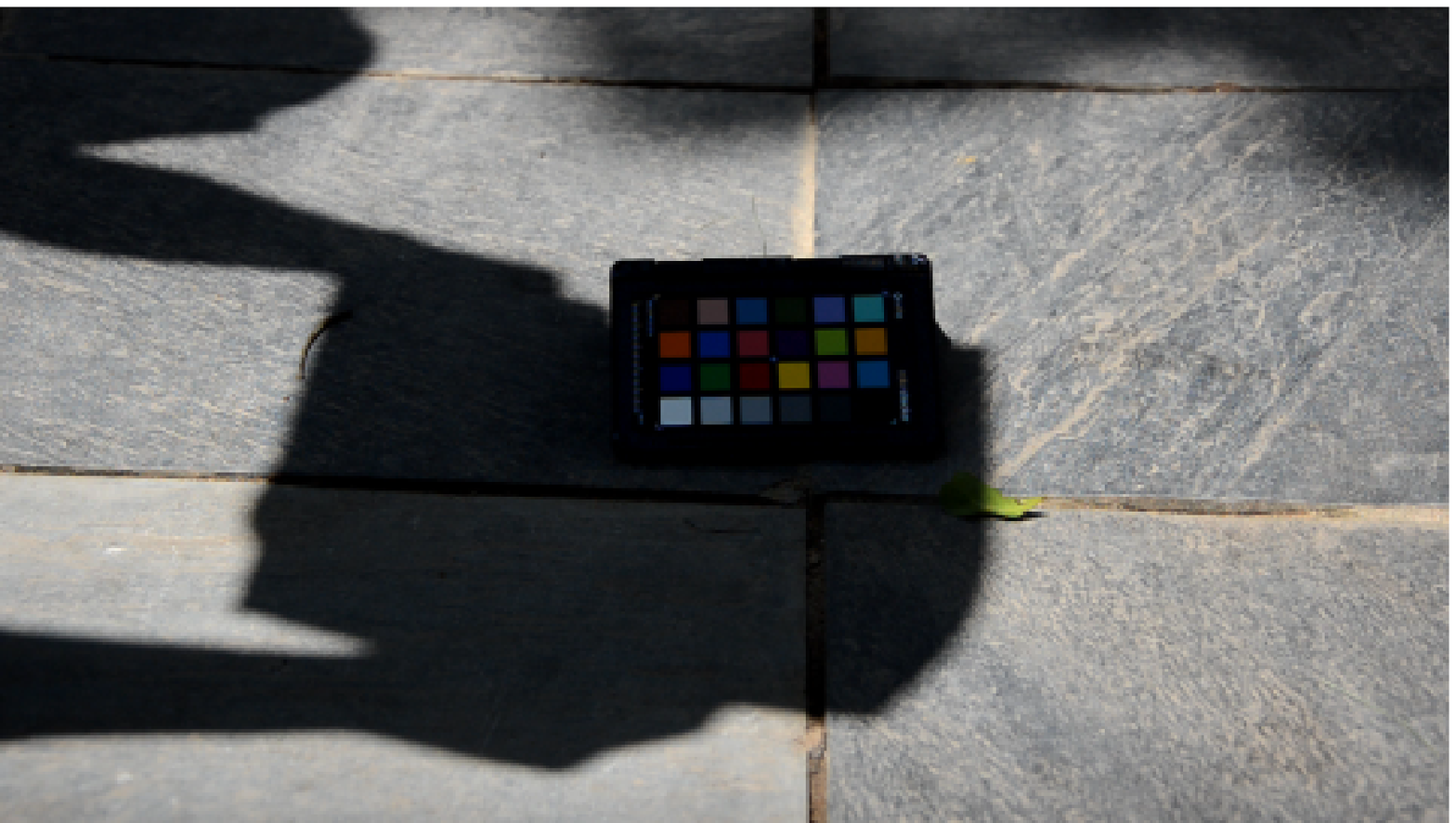}}
  \centerline{(b) $I_{\delta}$}\medskip 
\end{minipage}
\begin{minipage}[b]{0.31\linewidth}
  \centering
  \centerline{\includegraphics[height=1.3cm]{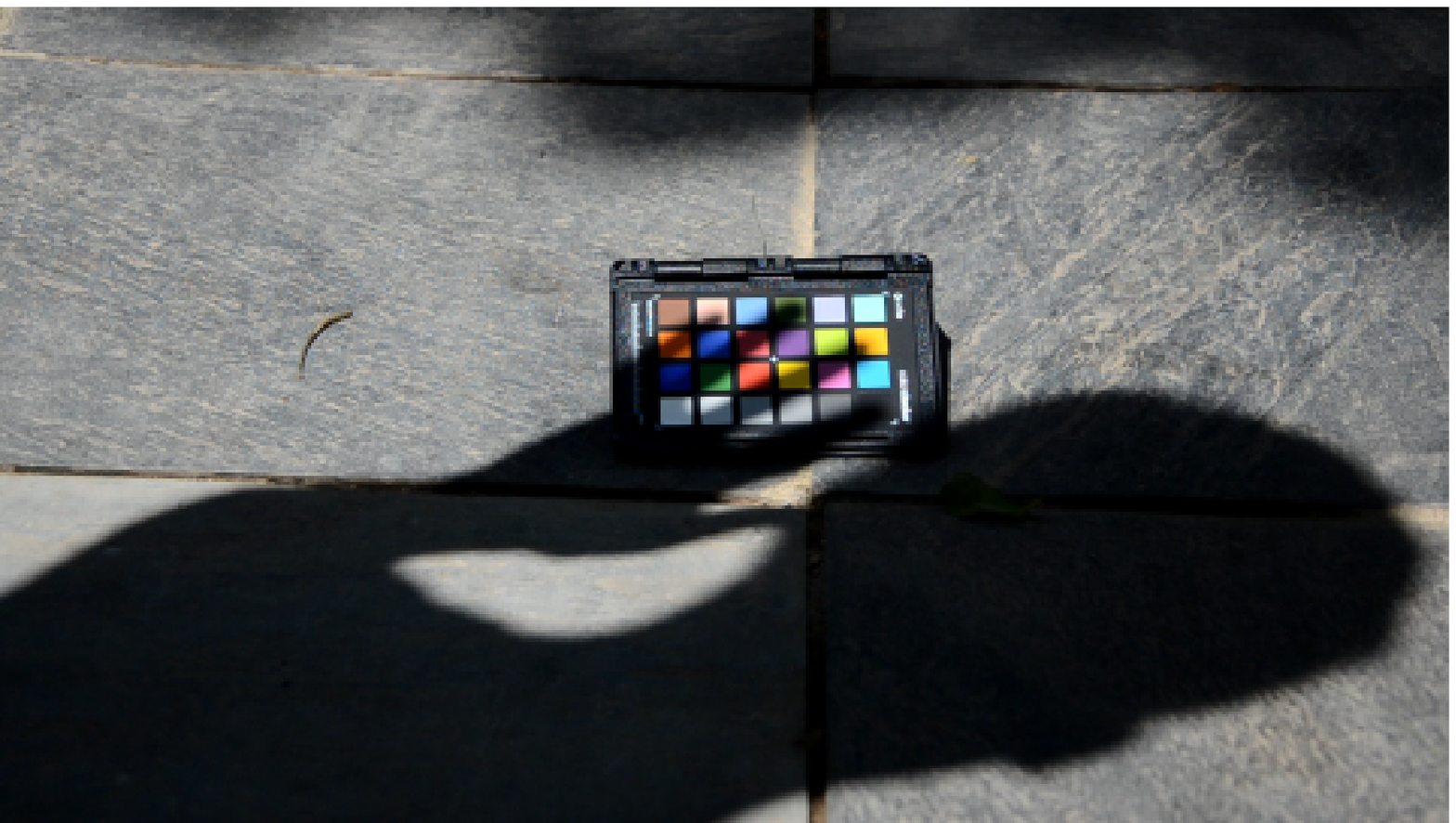}}
  \centerline{(c) $I_{\phi+\delta}$}\medskip 
\end{minipage}
\caption{A ColorChecker under different lighting conditions.} 
\label{fig:colorChecker}
\end{figure}

\begin{figure}

\begin{minipage}[b]{0.48\linewidth}
  \centering
  \centerline{\includegraphics[width=4.3cm]{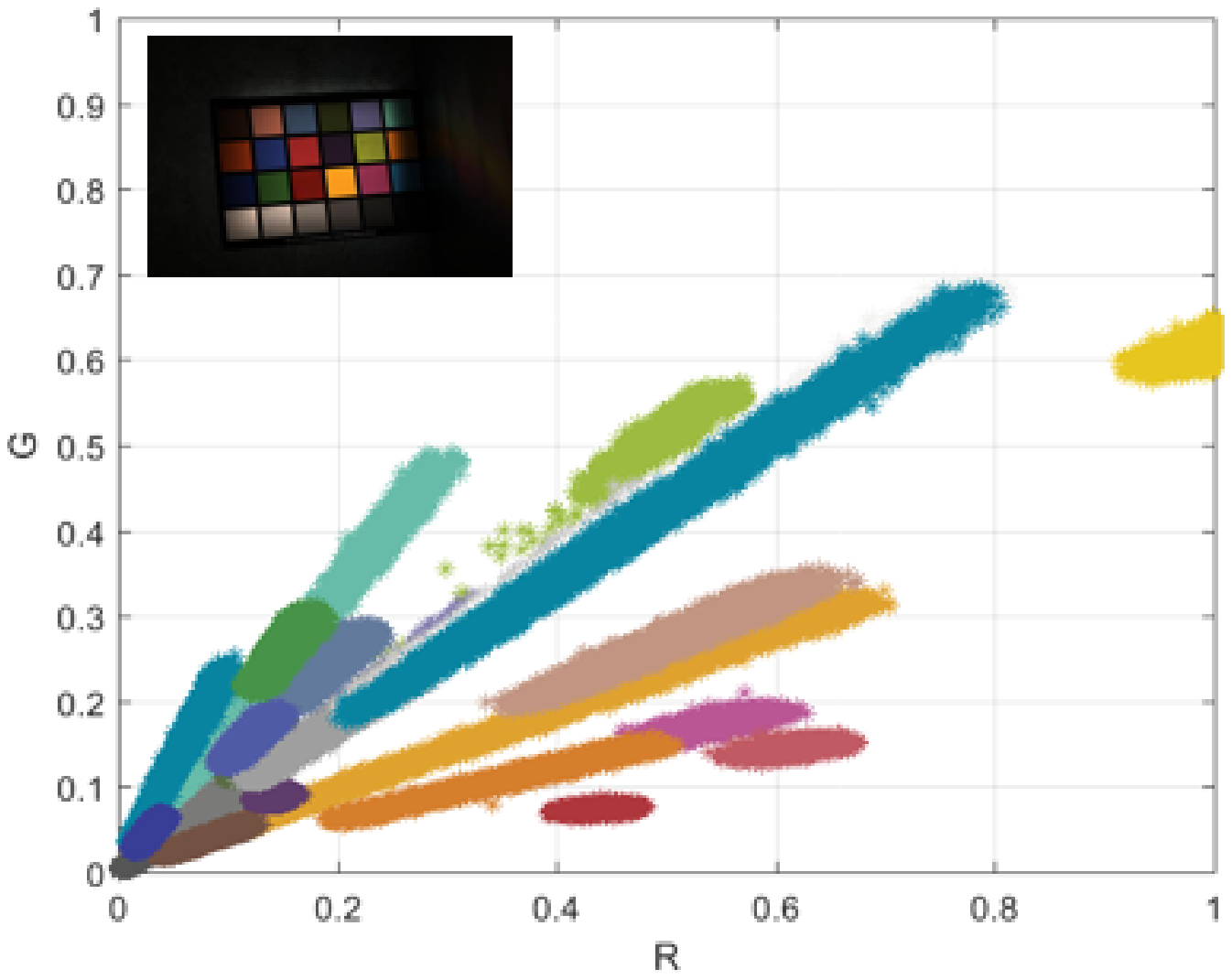}}
  \centerline{(a) $I_{\phi}$  }\medskip
\end{minipage}
\begin{minipage}[b]{0.51\linewidth}
  \centering
  \centerline{\includegraphics[width=4.3cm]{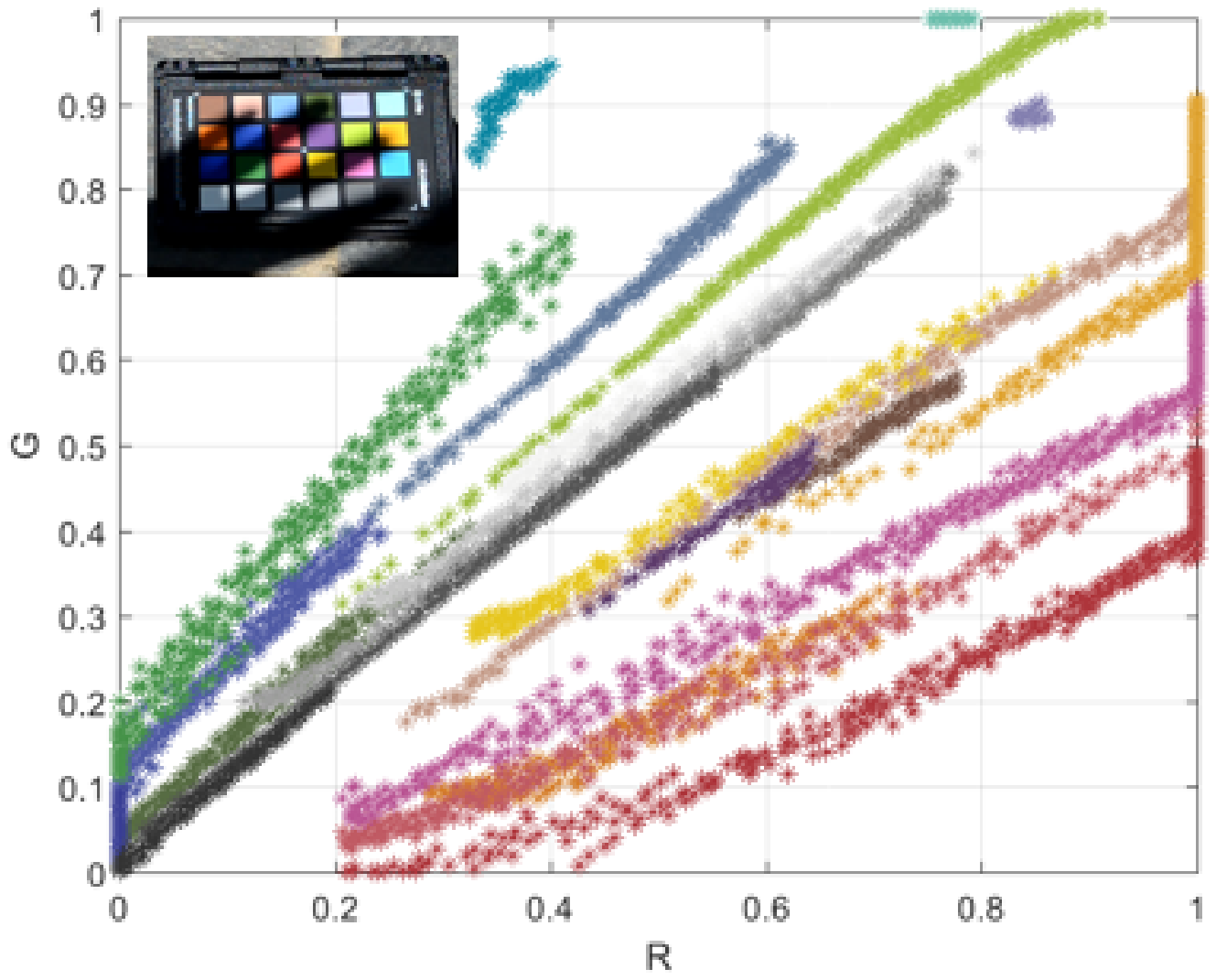}}
  \centerline{(b) $I_{\phi+\delta}$ }\medskip
\end{minipage}
\hfill
\begin{minipage}[b]{0.48\linewidth}
  \centering
  \centerline{\includegraphics[width=4.3cm]{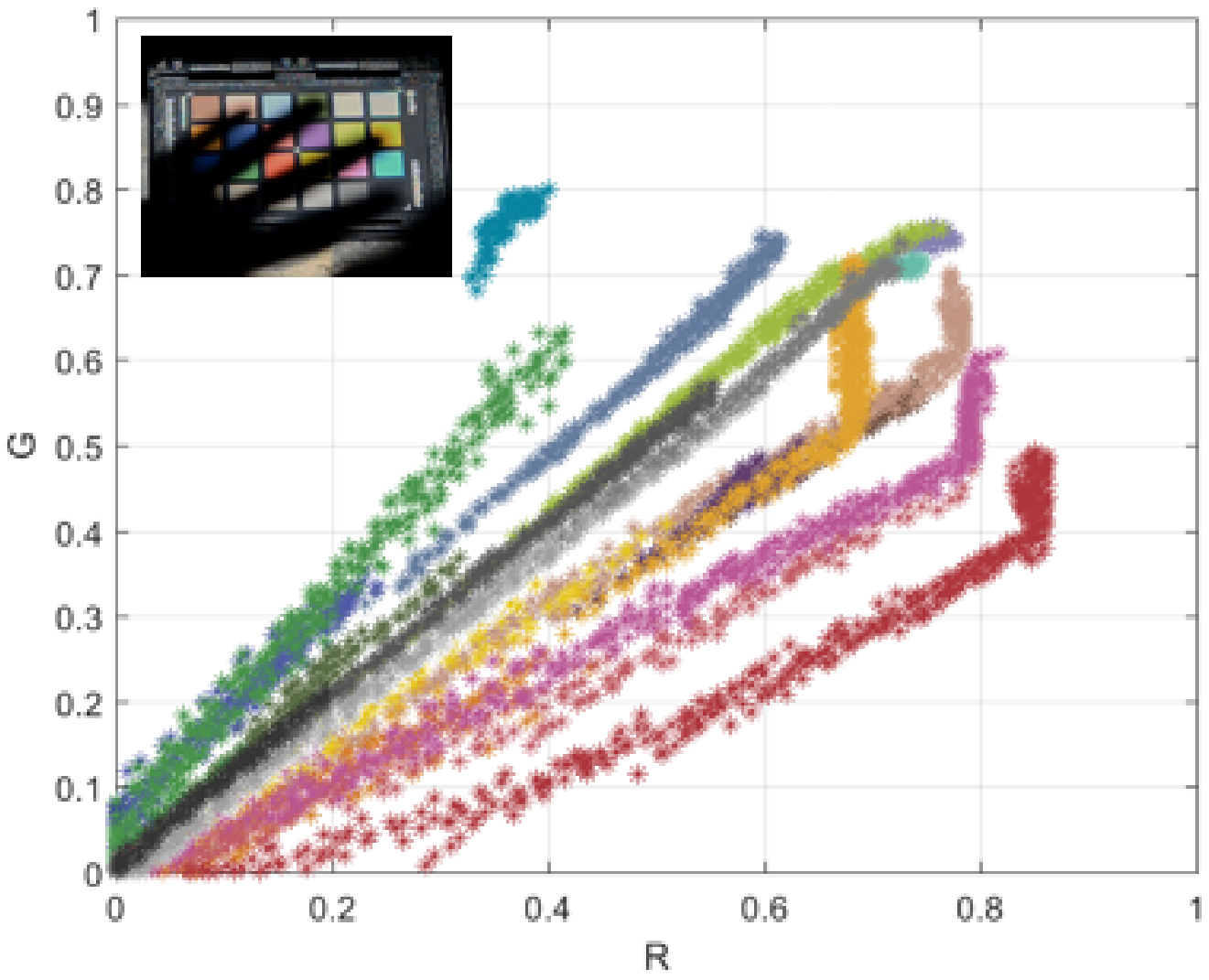}}
  \centerline{(c) $I_{\phi+\delta} - I_{\delta}$ }\medskip
\end{minipage}
\begin{minipage}[b]{0.50\linewidth}
  \centering
  \centerline{\includegraphics[width=4.3cm]{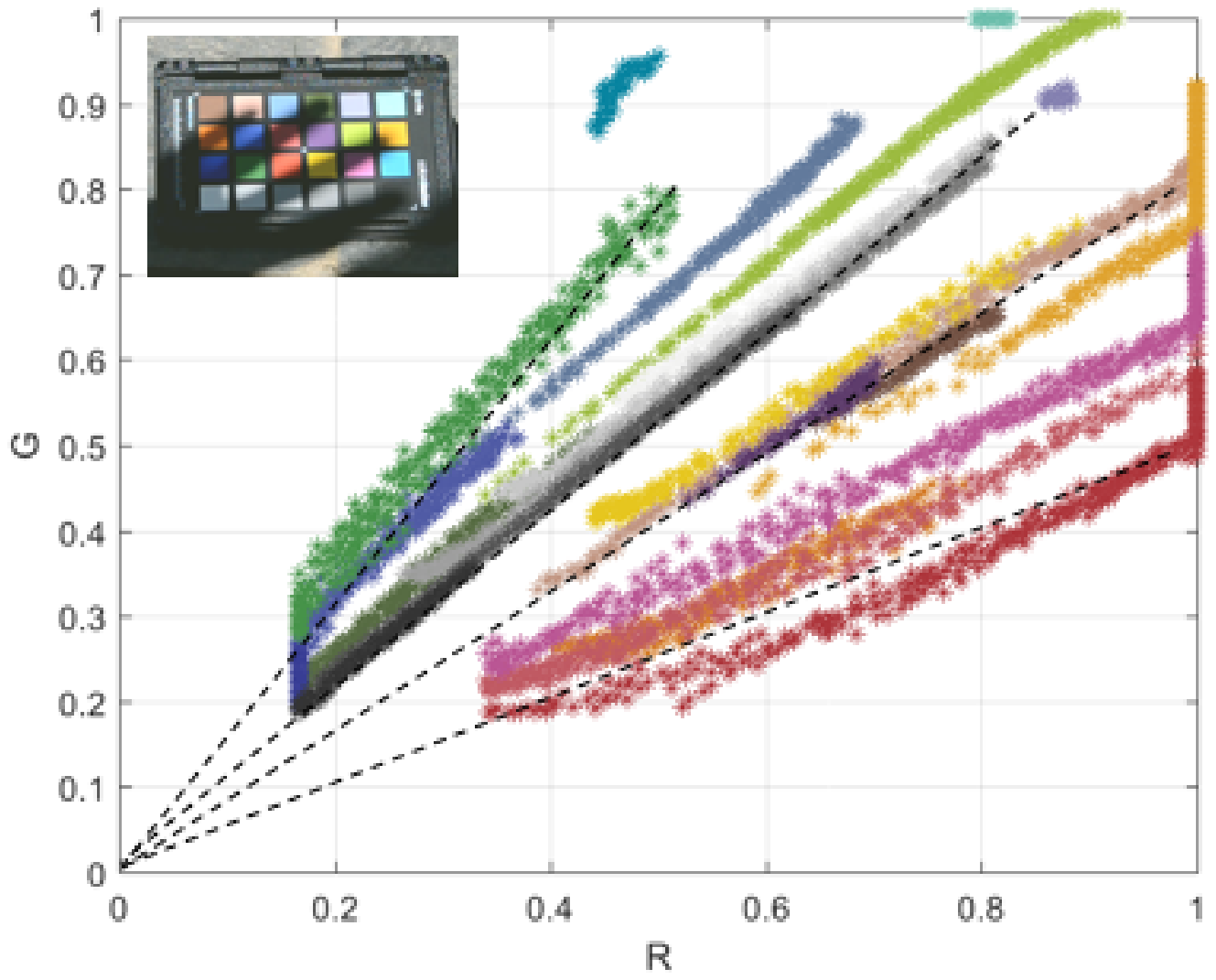}}
  \centerline{(d) $\tilde{I}_{\phi+\delta}$ }\medskip
\end{minipage}
\caption{Color distribution of different images. $RGB$ points are projected to $RG$ plane for better understanding. Each point is plotted with its corresponding color in Color Checker.
}
\label{fig:colorDistribution}
\end{figure}

To verify the proposed model, we took pictures of a ColorChecker exposed to different light conditions.
As shown in Fig.\ref{fig:colorChecker}, the ColorCheck located in image $I_\phi$ is only exposed to direct light
while that in $I_\delta$ and $I_{\phi+\delta}$ is fully and partly occluded under outdoor daylight respectively.
To observe the distribution of color lines,
pixels in 24 color areas of the ColorChecker are plotted into RGB space with its corresponding color, as shown in Fig.\ref{fig:colorDistribution}.

As for $I_\phi$, $\boldsymbol{\delta}$ of each pixel is a zero vector, and the sensor response is determined by $\boldsymbol{\phi}$. 
For the same material pixels captured by different color sensors,
their responses have a proportional relationship (cf. Eq.\ref{equ:phiIsLinear}). 
For same scenario captured by one camera, that proportion is only determined by the surface reflection property, thus pixels from the same material
surface will lie in one straight line passing the origin, as shown in Fig.\ref{fig:colorDistribution} (a).
%
\begin{equation}
\frac{\rho_i}{\rho_j} = \frac{\phi_i}{\phi_j} = \frac{ \int_\omega L^d(\lambda) S(\lambda) Q_i(\lambda) d\lambda }{ \int_\omega L^d(\lambda) S(\lambda) Q_j(\lambda) d\lambda}, 
\boldsymbol{\delta} = \boldsymbol{0}, i,j \in \{ 1,2,3 \}
\label{equ:phiIsLinear}
\end{equation}
%


As for $I_{\phi+\delta}$, we use hand to occlude the ColorChecker, 
creating various illumination conditions (umbra, penumbra, and lit areas).
Although pixels from the same material still distributed in a line, 
those lines do not intersect at origin due to the environment light,
as shown in Fig.\ref{fig:colorDistribution} (b).

As for $I_{\delta}$, the ColorChecker is entirely shrouded in shadow, 
thus $\boldsymbol{\phi}$ of each pixel is a zero vector.
Since $I_{\phi+\delta}$ and $I_{\delta}$ are taken in the same scene with the same camera settings,
we can remove the environment light part of $I_{\phi+\delta}$ by subtracting 
$I_{\delta}$.
As shown in Fig.\ref{fig:colorDistribution} (c), the offset is greatly attenuated in the resulting image $I_{\phi+\delta} - I_{\delta}$. 

%
In summary, 
Color consistency is established only in the case of no environment light. As the presence of environment light undermines the color consistency, lines consisted of pixels from the same material do not intersect the origin. By removing the contribution from environment light, those lines will intersect at the origin again.


%




\section{Offset Correction} 
\label{sec:offsetCorrection}

In most cases, the image of corresponding environment light $I_{\delta}$ is not available, 
so we need to find another way to perform offset correction. 
Noticing that the straight lines in $I_{\phi+\delta}$ are about to converge at one point, 
a simple and straight-forward way is to perform a linear transform.
Denote the location of that convergence point as $\boldsymbol{\varepsilon} = [\varepsilon_1, \varepsilon_2, \varepsilon_3]$, the offset-corrected 
sensor response $\boldsymbol{\tilde{\rho}}$ is defined as 
\begin{equation}
\boldsymbol{\tilde{\rho}} = \frac{\boldsymbol{\rho} - \boldsymbol{\varepsilon}}{\boldsymbol{1} - \boldsymbol{\varepsilon}}.
\end{equation}
%

With the help of ColorCheck, $\boldsymbol{\varepsilon}$ can be estimated easily.
As shown in Fig.\ref{fig:colorDistribution} (d), the offset of the resulting image is greatly attenuated with slight changes in appearance (a little bit brighter).
The remaining problem is to find $\boldsymbol{\varepsilon}$ for general images
without the help of ColorChecker.

After offset correction, 
the proportional relationship for responses of the same material pixels captured
by different color sensors
should be recovered.
Thus, $\tilde{\rho}_i / \tilde{\rho}_j$ is a material dependent constant.
We can use this characteristics to estimate $\boldsymbol{\varepsilon}$:
\begin{equation}
\forall i,j \in \{ 1,2,3 \},
\quad
\frac{\rho_i- \varepsilon_i}{\rho_j- \varepsilon_j} 
= \underbrace{
    \frac{\tilde{\rho}_i}{\tilde{\rho}_j} 
    \times \frac{1- \varepsilon_i} {1- \varepsilon_j} 
}_\textrm{Constant},
\end{equation} 
\begin{equation}
%
\Rightarrow
\underbrace{
  \sum_{i=1}^3 \frac{ \rho_i - \varepsilon_i}  {\rho_j - \varepsilon_j}
}_\textrm{Constant}  
=
\frac{ \sum_{i=1}^3 (\rho_i - \varepsilon_i) }  {\rho_j - \varepsilon_j}
\stackrel{ \rho_i \gg \varepsilon_i }{\approx }
\frac{\sum_{i=1}^3 \rho_i} {\rho_j - \varepsilon_j},
\end{equation} 
\begin{equation}
\rho_j \approx
\underbrace{
  \sum_{i=1}^3 \frac {\rho_j - \varepsilon_j} { \rho_i - \varepsilon_i }  
}_\textrm{Constant} 
\times \sum_{i=1}^3 \rho_i + \varepsilon_j.
\label{equ:linearRelationship}
\end{equation}
As implied in Eq.\ref{equ:linearRelationship}, 
there is an approximate linear relation between $\rho_j$ and $\sum_{i=1}^3 \rho_i$
among pixels from the same material.
Taking advantage of this relationship, we can calculate $\boldsymbol{\varepsilon}$ by following steps:
%

\begin{enumerate}[itemsep=0ex] 
  \item Manually select an area of interested material.
  \item Repeat 3-4 for each color channel ($j = 1,2,3$). 
  \item 
  Fit a straight line to approximate the relationship between $\rho_j$ and $\sum_{i=1}^3 \rho_i$.
  \item Take the intercept of the line as an estimation of $\varepsilon_j$.
\end{enumerate}







%

\begin{figure*}[t] 
\centering
\includegraphics[width=16cm]{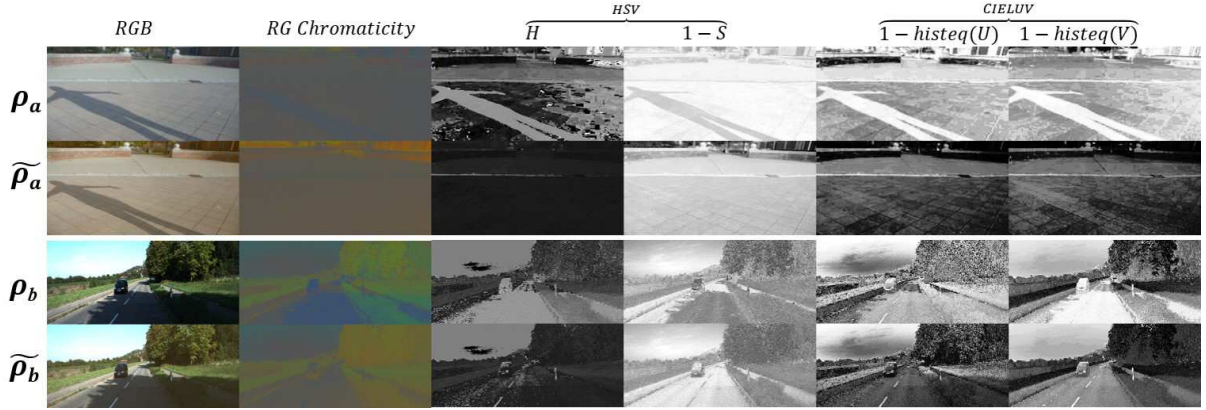}  
\caption{Color Space Conversion using $\boldsymbol{\rho}$ (raw) and $\boldsymbol{\tilde{\rho}}$ (offset-corrected). 
In order to make a clearer distinction, histogram equalization ( $histeq(.)$ ) is used to enhance some of the images.
In addition, 
some images are inverted ($I \gets 1 - I$) to look more natural.
Here $\boldsymbol{\tilde{\rho_a}}$ is from a shadow dataset \cite{gong2014interactive} and 
$\boldsymbol{\tilde{\rho_b}}$ is from a road dataset \cite{fritsch2013new}.
}
\label{fig:ColorSpaceConversion}
\end{figure*}

\section{Experimental Results}
\label{sec:exprimentalResults}

To illustrate the practical significance of offset correction, 
we conducted experiments from three aspects: color space conversion, 
image processing and analysis.

\textbf{Color Space Conversion.}
Given the importance of color processing in both Computer Vision and Graphics, color spaces abound.
While existing color spaces address a range of needs, none of them can free from the interference of severe shadows.
To verify the benefit of offset correction, 
we perform Color Space Conversion for original RGB images and offset-corrected RGB (ORGB) images.
Three common used color spaces are tested: 
RG Chromaticity, HSV and CIELUV. 
%
%
Experimental results show that 
the shadows in extracted color components
are greatly attenuated after offset correction, as shown in Fig.\ref{fig:ColorSpaceConversion}.



\textbf{Image Processing.}
As color components become more illumination-robust
after offset correction,
color-based image processing can be improved using ORGB instead of RGB.
Although there is no big difference between RGB and ORGB (c.f. Fig.\ref{fig-intro}(a)),
image processing results using them is quite different.
The segmentation results using color components show the color constancy has been recovered after offset-correction (c.f. Fig.\ref{fig-intro}(b)). 
Edge detection is free from the interference of severe shadow, 
and the object edges under shadow become clearer (cf. Fig.\ref{fig-intro}(c)).
As for feature match, the number of matched feature points is greatly increased (cf. Fig.\ref{fig-intro}(d)).
%

%

\textbf{Image Analysis.} 
Road detection is taken as an example to 
demonstrate the benefit of offset correction to image analysis. 
%
As a key technique of automatic driving, 
road detection algorithms suffer from the shadows on road surfaces \cite{BarHillel2012}.
The color-based road detection framework in \cite{ying2016mm} is employed to compare
the detection performances 
before and after offset correction.
Pixel-wise measurements are used to evaluate the performance of road detection,
including four measurements for quantitative evaluations (quality $\hat{g}$, detection rate $DR$, detection accuracy $DA$ and effectiveness $F$) 
and one qualitative measurement called valid road result index \textit{VRI} \cite{alvarez20103d}.
As shown in Table \ref{tbl:roadDetectionBenchmark}, the performance of 
road detection is improved after offset correction
in both quantitative and qualitative
measurements, especially in severe shadow cases.

\begin{table} 
\centering
\caption{Road detection performance on ROMA dataset\cite{Veit2008}.}
\tabcolsep=0.11cm
\begin{tabular}{|l|c|c|c|c|c|} 
\cline{2-6}
\multicolumn{1}{c|}{}                          & \multicolumn{5}{|c|}{Complete dataset} \\
\cline{2-6}
\multicolumn{1}{c|}{}                          & $\hat{g}$        & $DR$ & $DA$ & $F$ & $VRI$ \\ \hline
RGB 
& {.80 $\pm$ .23} & {.84 $\pm$ .22} & {.91 $\pm$ .20} & {.87 $\pm$ .20} & {81\%}  \\ \hline 
ORGB
& {.83 $\pm$ .18} & {.87 $\pm$ .19} & {.94 $\pm$ .14} & {.89 $\pm$ .15} & {84\%}  \\ \hline 

\multicolumn{1}{c|}{}                          & \multicolumn{5}{|c|}{
Severe shadow cases
} \\  
\cline{2-6}
\multicolumn{1}{c|}{}                          & $\hat{g}$ & $DR$ & $DA$ & $F$ & $VRI$ \\ \hline
RGB 
	     &{.75 $\pm$ .26} & {.79 $\pm$ .26} & {.88 $\pm$ .24} & {.82 $\pm$ .23} & {73\%}  \\ \hline 
ORGB
&{.83 $\pm$ .19} & {.87 $\pm$ .20} & {.93 $\pm$ .16} & {.89 $\pm$ .16} & {88\%}  \\ \hline

\end{tabular}
\label{tbl:roadDetectionBenchmark}
\end{table}

\section{Conclusions}


In this paper, 
we present an explanation of why existing techniques cannot perform well in severe shadow cases
though
theoretical deduction 
and experimental verification.
We attribute the reason to the offset caused by environment light, which is ignored in many literature.
Instead of modifying models or algorithms, 
we proposed an image pre-processing 
method to remove the offset. 
%
%
Experimental results show that offset correction can improve the performance of color space conversion in severe shadow cases. Besides, the proposed method can be applied to image processing and analysis
 via existing models or algorithms without any modification.
To encourage future works, we make the source code open,
as well as related materials. More testing
results can be found on our project website: https://baidut.github.io/ORGB/.



\vfill\pagebreak



\bibliographystyle{IEEEbib}

\end{document}